\documentclass[conference]{IEEEtran}
\IEEEoverridecommandlockouts
\usepackage{cite}
\usepackage{amsmath,amssymb,amsfonts}
\usepackage{algorithmic}
\usepackage{graphicx}
\usepackage{textcomp}
\usepackage{xcolor}
\usepackage{cite}
\usepackage{booktabs}
\usepackage{subcaption}
\usepackage{orcidlink}

\def\BibTeX{{\rm B\kern-.05em{\sc i\kern-.025em b}\kern-.08em
    T\kern-.1667em\lower.7ex\hbox{E}\kern-.125emX}}
\begin{document}

\title{C-PATH: Conversational Patient Assistance and Triage in Healthcare System}

\author{
\IEEEauthorblockN{Qi Shi\orcidlink{0000-0001-6530-0264}, Qiwei Han\orcidlink{0000-0002-6044-4530}}
\IEEEauthorblockA{School of Business and Economics \\Universidade Nova de Lisboa\\
Carcavelos, Portugal \\
\{qi.shi, qiwei.han\}@novasbe.pt}
\and
\IEEEauthorblockN{Cláudia Soares\orcidlink{0000-0003-3071-6627}}
\IEEEauthorblockA{School of Science and Technology \\ Universidade Nova de Lisboa\\
Caparica, Portugal \\
claudia.soares@fct.unl.pt}
}

\maketitle

\begin{abstract}
Navigating healthcare systems can be complex and overwhelming, creating barriers for patients seeking timely and appropriate medical attention. In this paper, we introduce C-PATH (Conversational Patient Assistance and Triage in Healthcare), a novel conversational AI system powered by large language models (LLMs) designed to assist patients in recognizing symptoms and recommending appropriate medical departments through natural, multi-turn dialogues. C-PATH is fine-tuned on medical knowledge, dialogue data, and clinical summaries using a multi-stage pipeline built on the LLaMA3 architecture. A core contribution of this work is a GPT-based data augmentation framework that transforms structured clinical knowledge from DDXPlus into lay-person-friendly conversations, allowing alignment with patient communication norms. We also implement a scalable conversation history management strategy to ensure long-range coherence. Evaluation with GPTScore demonstrates strong performance across dimensions such as clarity, informativeness, and recommendation accuracy. Quantitative benchmarks show that C-PATH achieves superior performance in GPT-rewritten conversational datasets, significantly outperforming domain-specific baselines. C-PATH represents a step forward in the development of user-centric, accessible, and accurate AI tools for digital health assistance and triage.
\end{abstract}

\begin{IEEEkeywords}
Large Language Models, Conversational AI, Patient Navigation, Medical Triage, Digital Health, Clinical Dialogue Systems
\end{IEEEkeywords}



\section{Introduction}
A patient navigator in the medical domain is defined as ``a barrier-focused intervention designed to help individual patients overcome obstacles to accessing and navigating the healthcare system promptly''~\cite{wells2008patient}. These roles include identifying barriers at the patient level, facilitating appointments, supporting communication between patients and providers, and providing health education~\cite{peart2018patient, mcbrien2018patient}. Despite their potential, patient navigation services face significant challenges due to variability in navigator qualifications, lack of standardization, and limited certification protocols~\cite{dixit2021navigating, ustjanauskas2016training, kokorelias2021factors}.

Recent advances in large language models (LLMs), such as ChatGPT, have prompted a growing interest in their application to medical contexts~\cite{omiye2024large, lee2024deep, tian2024opportunities}. LLMs have shown strong performance in clinical tasks, including question answering~\cite{kung2023performance}, summarization~\cite{van2024adapted}, and information extraction~\cite{lee2024deep}, and have shown potential to improve clinical reasoning~\cite{singhal2023large}. However, their integration in patient-facing medical conversations remains underdeveloped due to risks such as hallucination~\cite{chelli2024hallucination}, misinformation~\cite{kim2025medical}, and limited alignment of health literacy~\cite{clusmann2023future}. Additionally, the conversational fluency and trustworthiness of LLM outputs pose usability concerns, particularly in patient triage and navigation scenarios~\cite{jones2023artificial, thirunavukarasu2023large}.

In this paper, we introduce C-PATH (Conversational Patient Assistance and Triage in Healthcare), an LLM-based system designed to facilitate symptom recognition and triage patients to the appropriate medical departments through natural, multi-turn conversations. C-PATH uses patient-friendly terminology, allowing laypeople to describe symptoms effectively and understand AI responses clearly. Our approach consists of three key technical contributions:

\begin{enumerate}
    \item We design a multi-stage fine-tuning pipeline using an open-source LLaMA3 model, with stages aimed at acquisition of medical knowledge, understanding of dialogue, and summarization.
    \item We propose a novel dataset construction framework that converts structured differential diagnosis cases into realistic, multi-turn doctor-patient conversations using GPT prompting, allowing alignment with patient communication norms.
    \item We introduce a scalable multi-turn dialogue management module that prunes history and optionally summarizes earlier turns to stay within context limits while maintaining interaction coherence.
\end{enumerate}

Our experiments demonstrate that C-PATH not only produces coherent and accessible dialogues but also offers accurate department-level triage recommendations. Evaluations using GPTScore show improved understandability, informativeness, and specificity, while performance benchmarks indicate superior results compared to baselines. 

In general, C-PATH offers a promising step toward improving access and navigation to digital healthcare. By automating the initial check and triage of symptoms, our system has the potential to reduce wait times, optimize primary care doctor and specialist referrals~\cite{duarte2023dissecting,valdeira2023extreme}, and improve the efficiency of hospital workflows~\cite{han2018hybrid}. The automatic generation of EHR-compatible summaries further integrates LLMs into clinical documentation pipelines, creating value for both patients and providers.

\begin{figure*}[!htb]
  \centering
  \includegraphics[width=0.9\linewidth]{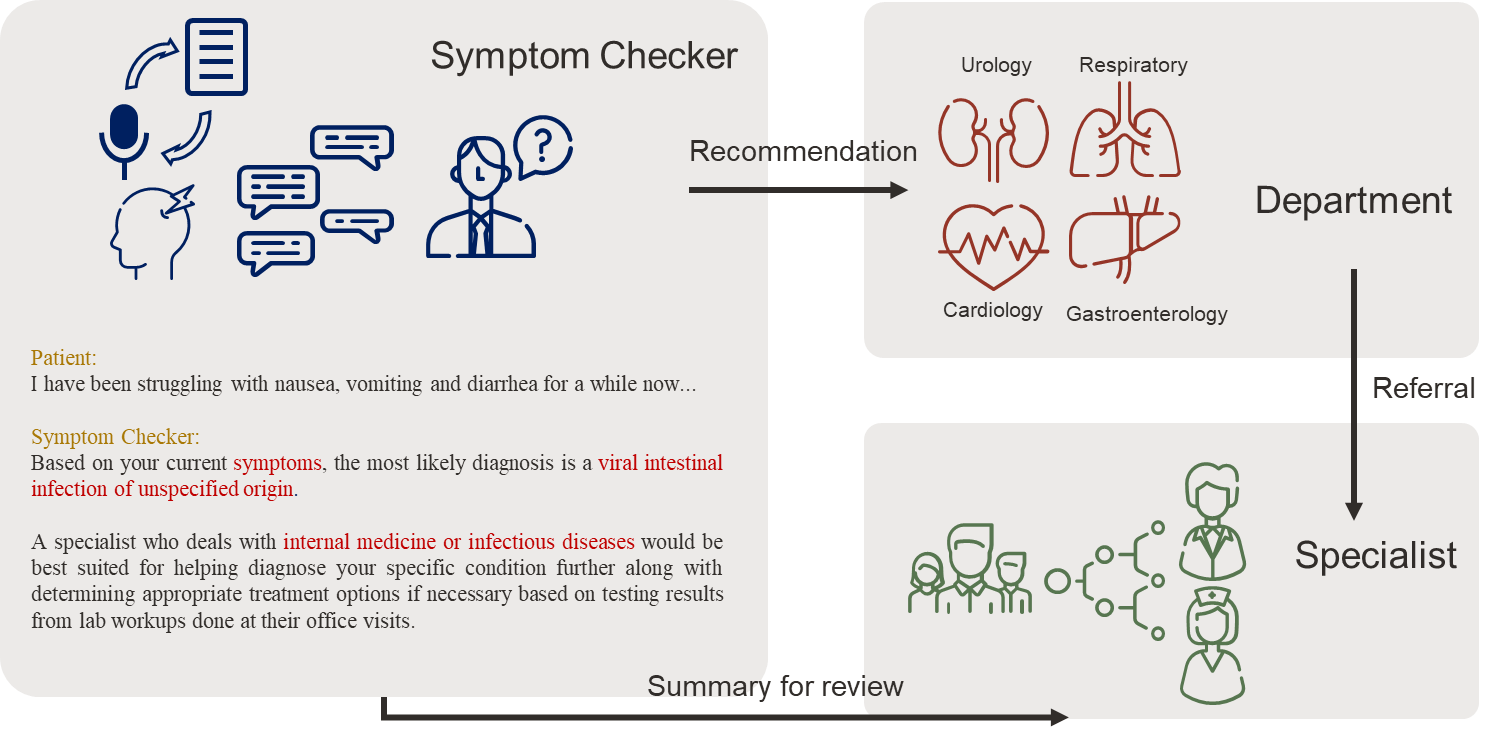}
  \caption{Overview of the C-PATH model framework. }
  \label{fig:framework}
\end{figure*}

\section{Related Work}
Symptom checkers for patient navigation have gained increasing attention as tools to improve the accuracy and efficiency of the diagnosis of medical conditions~\cite{wallace2022diagnostic}. Their development has been particularly important in public health crises, such as the COVID-19 pandemic, where timely monitoring and rapid response are essential. For example, \cite{munsch2020diagnostic} highlighted how the symptom checkers maintained public safety by tracking and reporting infection trends. Similarly, \cite{Zobel2023} demonstrated that platforms like Symptoma provided early warnings of COVID-19 spread, offering alternative data sources for policymakers and often anticipating official reports.

Another benefit of symptom checkers lies in user satisfaction, especially when integrated with self-triage and scheduling tools. \cite{Liu2022} found that such systems, when used during the pandemic, reduced unnecessary communication with healthcare providers by empowering users to schedule tests or consultations directly. This streamlining of patient flow has demonstrated significant operational value.

The accuracy of symptom checkers is also essential for positive health outcomes. For example, \cite{peven2023assessment} evaluated Flo Health's symptom checkers to identify conditions such as endometriosis, uterine fibroids, and polycystic ovary syndrome. They reported high diagnostic accuracy, suggesting that these tools could reduce time to diagnosis and improve healthcare outcomes. In a different domain, \cite{maturana2022advances} showed that integrating AI into microscopy imaging improved malaria diagnosis in low-resource settings, demonstrating a greater applicability of AI-powered diagnostic tools.

However, symptoms checkers show notable limitations in emergency care. \cite{abensur2023exploratory} compared them to emergency physicians using Objective Structured Clinical Examinations and found that human experts significantly outperformed symptom checkers in both primary and secondary diagnoses. This highlights the need to limit such tools to supportive rather than definitive roles in emergency contexts~\cite{fraser2017limitations}.

Consequently, AI-powered systems are better positioned to support healthcare workflows than to replace clinical decision-making. \cite{hammoud2024evaluating} investigated how AI improves automation in digital healthcare systems, finding benefits in operational efficiency and patient outcomes. However, \cite{you2023user} noted that current symptom checkers often do not meet expectations for conversational design, input flexibility, and language clarity, indicating the need for more user-friendly AI interfaces.

Recent progress in large language models (LLMs) provides a pathway to address these design shortcomings. Proprietary models such as Google's Med-PaLM 2 and MedLM~\cite{singhal2023large}, and open-source alternatives such as HuaTuo~\cite{wang2025knowledge} and ChatDoctor~\cite{li2023chatdoctor}, have shown promise. HuaTuo, for example, incorporates structured and unstructured medical knowledge from the Chinese Medical Knowledge Graph (CMeKG) to improve domain specificity. ChatDoctor leverages on-line medical dialogues for fine-tuning, producing more contextually relevant responses.

Despite these advancements, significant challenges remain for the deployment of LLMs in real-world healthcare applications. These include comprehensive medical knowledge~\cite{li2025merging}, managing patient medical history~\cite{goodman2024ai}, accurately recording conversations~\cite{adedeji2024sound}, and designing intuitive interfaces~\cite{sun2024trust}. Overcoming these issues is vital for the effective integration of LLM-based conversational systems into hospital infrastructure, with the goal of enhancing, not replacing, professional healthcare services~\cite{wen2024leveraging}.

\newpage
\section{Methodology}
Figure~\ref{fig:framework} provides an overview of the interaction flow and functionality of the model within the healthcare system. Initially, the component of the symptoms checker interacts with the patient to gather information about the symptoms through multi-turn conversations and perform a preliminary diagnosis. Based on this information, it recommends the most appropriate medical department for the patient's condition. Subsequently, the patient is referred to the relevant specialist. Throughout this process, the model also generates a concise summary of the interaction as Electronic Health Records (EHR), which can be reviewed by the specialist to inform further clinical decision-making and treatment.

\subsection{Training Workflow}
\subsubsection{Knowledge Injection}
The LMFlow toolkit is used throughout the training and fine-tuning workflows. LMFlow is an open-source large language model training and fine-tuning framework designed to support mainstream open-source models~\cite{diao-etal-2024-lmflow}. It natively integrates efficient parameter fine-tuning technologies, such as LoRA, significantly reducing training costs and allowing rapid customization and optimization of large-scale models.

\begin{figure}
    \centering
    \includegraphics[width=1\linewidth]{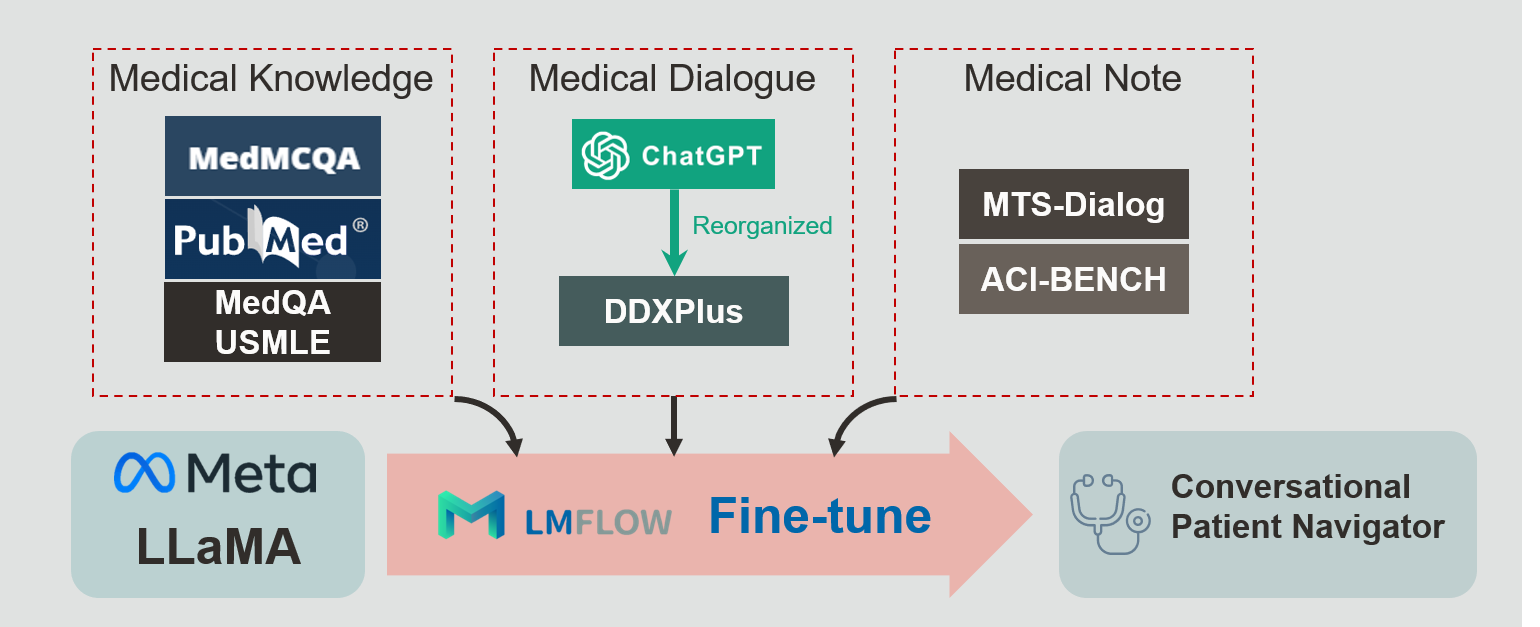}
    \caption{Training workflow for C-PATH.}
    \label{fig:training}
\end{figure}

The model employed in this study is based on LLaMA3 and is trained in a three-stage process (see Figure~\ref{fig:training}). In the first training stage, we focus on injecting extensive medical knowledge into the model. This step involves leveraging three widely used medical question-answering datasets: PubmedQA~\cite{jin2019pubmedqa}, MedQA-USMLE~\cite{jin2021disease}, and MedMCQA~\cite{pmlr-v174-pal22a}. These datasets encompass diverse and comprehensive content, ranging from clinical case questions to medical examination preparation materials. They are crucial to equipping the model with a broad and deep understanding of medical concepts, terminologies, symptoms, and related medical information.

\subsubsection{Instruction Tuning}
After acquiring the foundational medical knowledge, the model proceeds to the second training stage, emphasizing fine-tuning medical conversation skills. In this phase, we use a doctor-patient conversation dataset designed to simulate realistic medical interactions and consultations~\cite{FansiTchango2022}. The objective here is to enhance the model's capability to engage effectively in multi-turn dialogues. This fine-tuning ensures that the model can accurately interpret and respond contextually to complex, prolonged medical conversations. Such improvements are critical for delivering medically relevant, coherent, and contextually appropriate conversational responses.

To enable such multi-turn interactions in a reliable and scalable manner, we introduce a dialogue history management mechanism to maintain coherence, manage token limits, and preserve conversational flow.

\subsubsection{Multi-Turn Dialogue and History Management}
Our system is designed to support multi-turn conversations between patients and the C-PATH. This capability is essential for simulating realistic medical assistance scenarios, where symptoms may be revealed progressively and clarification questions are necessary for accurate recommendation.

To ensure scalability and coherence, we adopt the following approach for conversation history management.

\begin{itemize}
  \item \textbf{Context Window Pruning}: We apply a sliding window mechanism where only the most recent \textit{N} dialogue turns are retained in the prompt context. This ensures that the total token length remains within the model’s maximum input length, while preserving the most relevant recent context.

  \item \textbf{Turn-Level Summarization}: For longer sessions, we optionally condense earlier turns into a system-level summary using a lightweight LLM-based summarizer before appending new conversational turns. This allows the assistant to retain essential background knowledge without exceeding token limits.

  \item \textbf{Speaker Tagging and Role Preservation}: Each conversational turn is explicitly tagged with speaker roles (e.g., [Patient], [Assistant]) to maintain turn alignment, preserve conversational roles, and prevent instruction leakage or loss of speaker identity across turns.
\end{itemize}

This history management framework is critical to enabling the model to track evolving symptom narratives, follow up with clarifying questions, and iteratively refine department recommendations over multiple turns of patient interaction. It also contributes to consistency in clinical logic and fluency in longer sessions.

\subsubsection{Summarization Tuning}

The third and final stage of the training workflow involves using summaries learned from summary notes of doctor-patient conversation scenarios~\cite{mts-dialog,yim2023aci}. This step aims to transform the detailed conversations facilitated by the C-PATH into summarized clinical notes. These notes serve as references for medical specialists and are systematically integrated into Electronic Health Records (EHR), thus enhancing continuity of care and providing efficient documentation of patient interactions.

\subsection{Model}

\textbf{LLaMA3}: We selected LLaMA3-8B as the foundation model due to its strong performance on biomedical question-answering benchmarks such as PubMedQA. As shown in previous work~\cite{diao-etal-2024-lmflow}, a fully fine-tuned LLaMA3-8B outperformed larger models like ChatGPT in domain-specific tasks. Compared to models such as GPT-J or BLOOM, LLaMA offers a better balance between performance, model size, and computational efficiency, which is critical for future deployment in resource-constrained clinical environments, such as mobile platforms. Its open-source accessibility also provides full control over model customization, an essential feature in privacy-sensitive medical applications.

\subsection{Data}

\subsubsection{Medical Knowledge Datasets} Three medical question-answering dataset are included in the fine-tuning process to inject medical domain knowledge into LLM as follows:

\textbf{PubmedQA} is a question-answering dataset based on the PubMed database, designed for biomedical applications~\cite{jin2019pubmedqa}. It contains 1,000 expert-annotated QA instances, 612,000 unlabeled QA instances, and approximately 211,300 artificially generated QA instances.

\textbf{MedQA-USMLE} is a question-answering dataset specifically tailored for U.S. medical students and physicians preparing for the US Medical Licensing Examination~\cite{jin2021disease}. It comprises 12,723 questions covering extensive medical knowledge relevant to clinical practice.

\textbf{MedMCQA} is a large-scale multiple-choice question-answer dataset containing approximately 194,000 records~\cite{pmlr-v174-pal22a}. It covers over 2,400 healthcare topics across 21 distinct medical subjects, serving as a valuable resource for medical entrance exam preparation and comprehensive medical training.

\subsubsection{Medical Dialogue Datasets}

Doctor-patient dialogue scenarios typically occur in outpatient clinical settings, making these data challenging to collect. Existing open-source datasets, such as MedDialog~\cite{zeng-etal-2020-meddialog}, usually contain single-take conversations that focus primarily on diagnoses and treatment recommendations. This structure significantly differs from the conversational objectives of this paper, which aims to facilitate natural, multi-turn interactions between patients and medical guidance systems. To address these limitations and align more closely with the requirements, without loss of generality, we used a random sample of 5000 conversations from the DDXPlus dataset (\texttt{data\_5k\_ddxplus}).

\textbf{DDXPlus} is a comprehensive dataset comprising approximately 1.3 million patient cases, each containing differential diagnoses, confirmed pathologies, detailed descriptions of symptoms, and patient medical history~\cite{FansiTchango2022}. Unlike traditional datasets that typically contain binary symptom indicators, DDXPlus provides categorical and multi-select symptom data organized hierarchically, which significantly facilitates logical and interactive patient conversations.

\begin{figure}[!htb]
    \centering
    \includegraphics[width=1\linewidth]{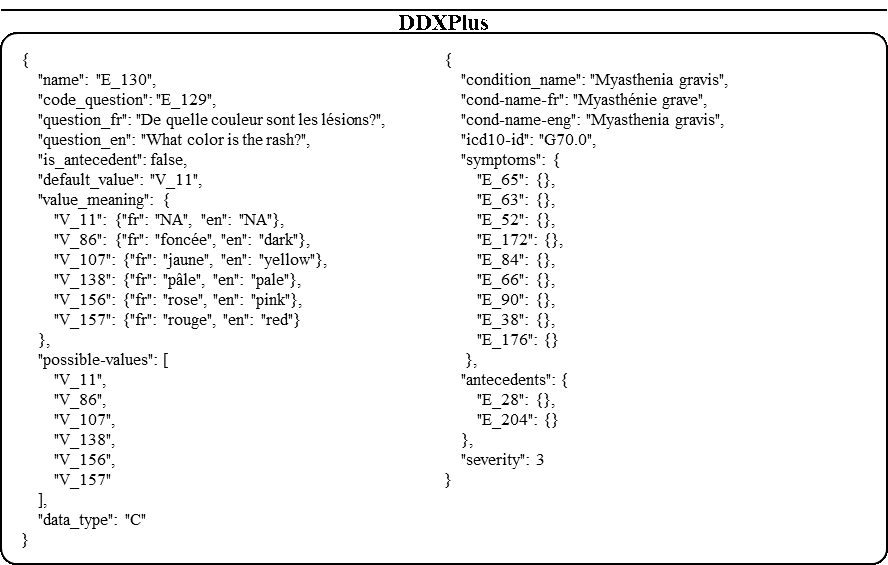}
    \caption{Example of evidence and pathology description in DDXPlus. }
    \label{fig:ddxplus_example}
\end{figure}

Although the DDXPlus dataset does not directly include conversational dialogues, we use its detailed symptom descriptions to construct artificial doctor-patient conversations. Initially, as Figure \ref{fig:ddxplus_example} shows, DDXPlus symptom-related questions often contained complex medical terminology, making them potentially difficult for patients to understand. To ensure clarity and promote natural human-like interactions, these queries were carefully rewritten using patient-friendly language. Medical terms were specifically avoided during the rewriting process, reflecting real clinical scenarios in which physicians typically use layman’s terms to facilitate patient understanding. For example, the term ``iliac wing(R)'' was rewritten into easily comprehensible phrases such as ``right side of my lower back" or ``near my right hip.''

\begin{figure}
    \centering
    \includegraphics[width=1\linewidth]{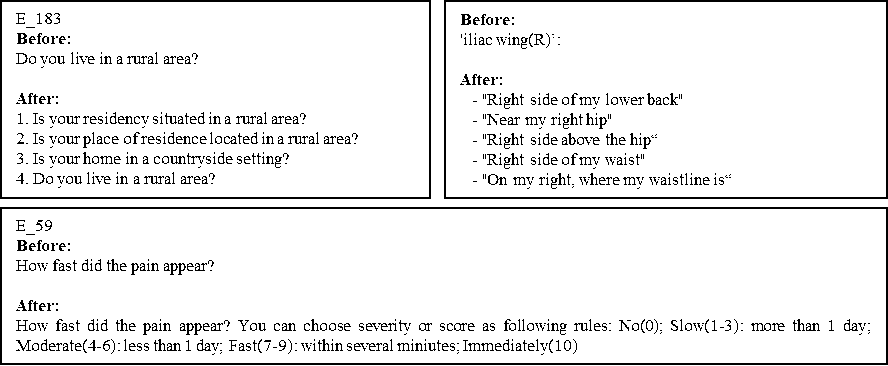}
    \caption{Example of term rewriting and question optimization. }
    \label{fig:rewrite}
\end{figure}

To further enhance the realism and diversity of these conversations, each symptom-related question was reformulated into multiple variants using GPT-3.5 Turbo. For each question, as illustrated by \texttt{E\_183} in Figure~\ref{fig:rewrite}, we generated four different versions to ensure variety. Questions about symptom severity, such as those illustrated by \texttt{E\_59}, were also rewritten to provide multiple intuitive approaches to understanding severity rather than relying solely on numeric scales (1-10). Similarly, affirmative patient responses were diversified through commonly used conversational expressions generated by GPT-3.5 Turbo, such as ``I think so," ``Absolutely," ``Of course," ``Definitely," and ``For sure," rather than the simpler ``Yes." These reformulated questions and responses were compiled into an artificial dataset, termed \texttt{data\_5k\_artificial}. Furthermore, we used GPT-3.5 Turbo to completely rewrite the entire original DDXPlus-based conversation dataset, creating \texttt{data\_5k\_GPT}. This allowed us to directly compare the performance and effectiveness of GPT-generated conversations with manually revised data. While human rephrasing introduces richer linguistic variety, the GPT-3.5 rewriting consistently outperforms in downstream tasks due to its greater stylistic consistency, grammatical coherence, and normalized response structure. These qualities ensure smoother conversational flow, reduced ambiguity, and more uniform input formats, which are especially beneficial for model training and inference. Table \ref{tab:dataset} provides an overview of the three datasets used for evaluation.

\begin{table}[!htb]
\caption{Description of conversational datasets used for training and evaluation.}
\label{tab:dataset}
\begin{tabular}{lp{5cm}}
\toprule
\textbf{Dataset} & \textbf{Description} \\
\midrule
\texttt{data\_5k\_ddxplus} & 5k multi-turn conversation samples directly composed of raw questions and answers from DDXPlus\\
\texttt{data\_5k\_artificial} & 5k multi-turn conversation samples composed of manually rewritten questions and answers\\
\texttt{data\_5k\_GPT} & 5k multi-turn conversation samples entirely rewritten by GPT-3.5 Turbo\\
\bottomrule
\end{tabular}
\end{table}

\subsubsection{Medical Conversation Notes Datasets}
Typically, hospitals do not capture complete transcripts of doctor-patient conversations; instead, they record interactions as fragmented highlights or concise summaries in medical records. Two medical conversation notes datasets are employed to fine-tune the C-PATH for summarized doctor-patient conversations for specialist review:

\textbf{MTS-Dialog} dataset comprises approximately 1,700 short doctor-patient conversations accompanied by corresponding summaries~\cite{mts-dialog}. Additionally, this dataset includes an augmented subset containing around 3,600 conversation-summary pairs. This augmented subset was created by back-translating the original 1,200 training pairs into French and Spanish, thereby increasing the dataset's diversity and comprehensiveness.

\textbf{ACI-BENCH} dataset features complete doctor-patient conversations paired with detailed clinical notes. It incorporates data splits utilized in the MEDIQA-CHAT 2023 and MEDIQA-SUM 2023 challenges, providing robust benchmarks for clinical dialogue summarization tasks~\cite{yim2023aci}.

Given that these summary records frequently lack coherence, detail and completeness, to address these limitations, we further used GPT-3.5 Turbo to generate structured summaries from our designed conversations, specifically from the data set \texttt{data\_5k\_ddxplus}. Compared to traditional templated summarization methods, GPT-3.5 Turbo can more easily adapt to specific conversational contexts, producing summaries that are logical, professionally formatted, and thoroughly organized, thus significantly improving the quality of clinical documentation.

\subsection{Fine-tuning Framework}

To enhance the efficacy of generalist LLMs for medical applications, researchers have recognized the importance of training and/or fine-tuning these models on large, in-domain datasets~\cite{peng2023study}. To address this need, we integrated the LMFlow toolkit into the fine-tuning workflow~\cite{Diao2023}, which simplifies the fine-tuning and inference processes of LLMs. It can incorporate instruction tuning methodologies designed to improve LLMs by explicitly training them to follow natural language instructions and commands. The fine-tuning process for our model was performed on two Nvidia A10 graphics cards, employing the hyperparameters listed in Table \ref{tab:hyperparams}.

\begin{table}[!htb]
\caption{Hyperparameters used in fine-tuning the C-PATH model. }
\label{tab:hyperparams}
\begin{tabular}{p{3.8cm}p{4.5cm}}
\toprule
\textbf{Hyperparameter} & \textbf{Description}\\
\midrule
$num\_train\_epochs: 2$ & Number of epochs used for training the model on the collected dataset.\\
$learning\_rate: 2e-5$ & Learning rate used during model fine-tuning.\\
$block\_size: 128$ & Maximum sequence length for tokenized inputs; datasets are truncated into segments of 128 tokens for training.\\
$per\_device\_batch\_size: 6$ & Batch size per GPU used during fine-tuning.\\
$use\_lora: yes$ & Indicates whether Low-Rank Adaptation (LoRA) is employed.\\
$lora\_r: 8$ & Rank parameter for LoRA method.\\
$bf16$ & Specifies mixed precision mode as BF16 (brain floating point 16), providing enhanced numerical precision compared to FP16.\\
$dataloader\_num\_workers: 1$ & Number of processes used for preprocessing during training.\\
\bottomrule
\end{tabular}
\end{table}

\textbf{Conversation Tuning} is a specialized form of instruction tuning that enables LLMs to effectively handle conversational interactions, extending beyond simple text completion tasks. It demands that models not only understand context, but also manage multi-turn dialogues. Due to these specific requirements, conversation tuning requires more precise formatting compared to traditional text fine-tuning. Improper formatting can lead to performance errors even in adequately trained models. A crucial aspect distinguishing conversation tuning from general text fine-tuning is the use of special end-of-turn markers. Examples of such markers include \texttt{</s>} used in Vicuna and \texttt{<|im\_end|>} employed in OpenAI's ChatML. These markers explicitly indicate the end of each dialogue turn, enabling the model to determine when a suitable response is complete.

\begin{figure}[!h]
\centering
\includegraphics[width=\linewidth]{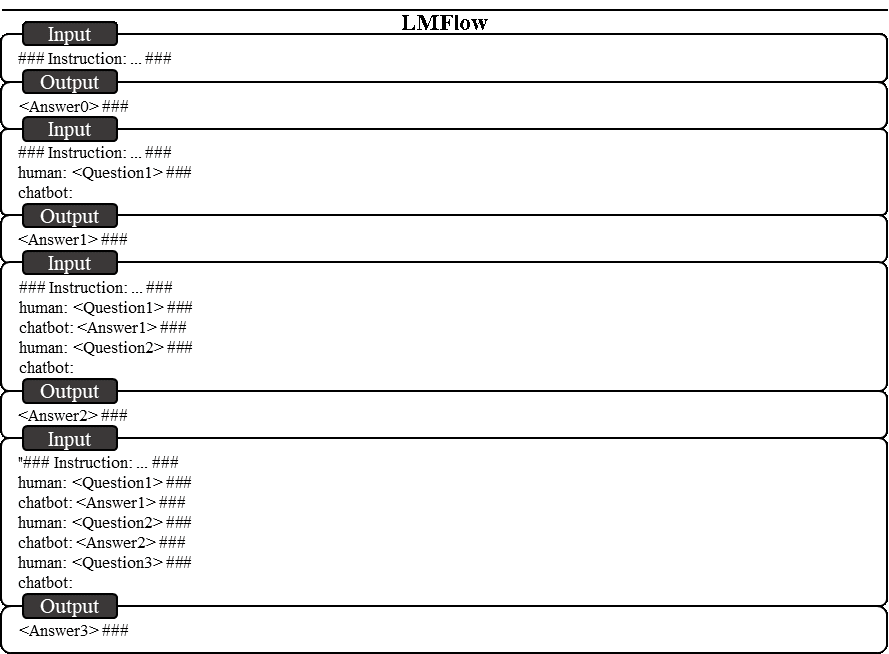}
\caption{Conversation formatting structure for fine-tuning with LMFlow.}
\label{fig:conversation-structure}
\end{figure}

We adopted the conversational turning strategy to structure our conversational prompts dataset into multi-turn conversations. We used the marker \texttt{\#\#\#} to indicate the end of each conversational exchange. As Figure~\ref{fig:conversation-structure} shows, the conversations were systematically decomposed into multiple individual dialogue turns, adding each turn sequentially, ensuring the complete preservation of contextual information.

\subsection{Model Evaluation}

To assess the potential performance of our model, we evaluated the quality of the DDXPlus dataset, previously introduced, as it is essential for enabling effective multi-turn dialogue interactions. To systematically assess the quality of the dialogue, we use GPTScore to assess the quality of the conversation in six dimensions (Specificity, Flexibility, Understandability, Informativeness, Patience and Accuracy)~\cite{fu-etal-2024-gptscore}. Unlike traditional NLG metrics such as BLEU or ROUGE—which are limited to surface-level lexical overlap—GPTScore allows for instruction-based, semantics-aware evaluation of generated text using natural language prompts. This makes it particularly well-suited for evaluating multi-turn, patient-facing medical conversations where nuance, tone, and clarity are critical. It also enables a more interpretable multidimensional assessment compared to black-box automatic metrics. Specifically, we evaluated different datasets derived from DDXPlus in six different aspects, as detailed in Table~\ref{tab:gptscore_aspects}.

\begin{table}[!htb]
\centering
\begin{tabular}{lp{7cm}}
\toprule
\textbf{Aspect}& \textbf{Definition}\\
\midrule
SPE & Are the responses of the Conversational Patient Navigator specific enough to the context rather than generic?\\
FLE & Is the Conversational Patient Navigator flexible and adaptive to individual patient interests and responses?\\
UND & Does the Conversational Patient Navigator clearly communicate information in a way easily understood by patients?\\
INF & Do the Conversational Patient Navigator's questions effectively gather sufficient information to provide accurate recommendations?\\
PAT & Do the questions from the Conversational Patient Navigator potentially lead to patient impatience?\\
ACC & Is the specialist recommended by the Conversational Patient Navigator accurately aligned with the patient's needs?\\
\bottomrule
\end{tabular}
\caption{Definitions of evaluation aspects in GPTScore.}
\label{tab:gptscore_aspects}
\end{table}

Since our model's primary objective is to accurately direct patients to the appropriate medical departments, the accuracy of recommendations on the validation set is the critical metric for our evaluation. To benchmark this, we selected BERT, a widely recognized model also based on the Transformer architecture, as our baseline for comparison. In our evaluation, the entire conversation serves as input, with departmental recommendations as targets. Each dataset was partitioned into 70\% training, 20\% testing, and 10\% validation subsets to ensure a thorough and robust evaluation process.

\section{Results}
\subsection{Exploratory Data Analysis}

\begin{figure}
    \centering
    \includegraphics[width=\linewidth]{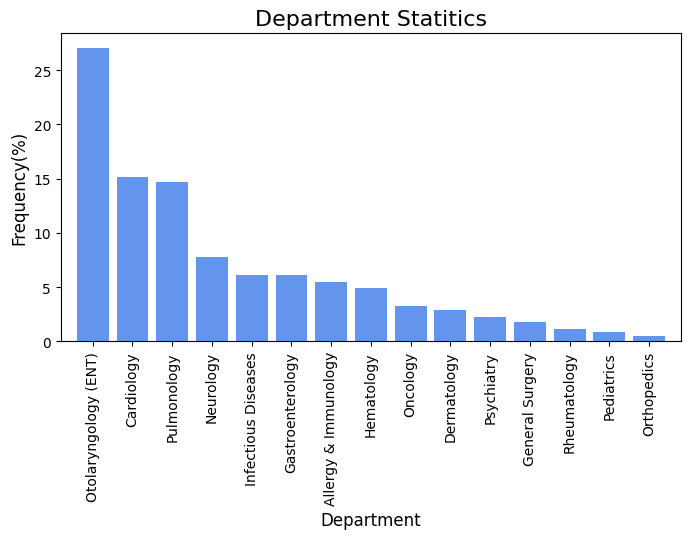}
    \caption{Department frequency distribution in the $data\_5k\_ddxplus$ dataset. }
    \label{fig:department_freq}
\end{figure}

Our data sampling strategy preserves the disease distribution in the original DDXPlus dataset by assigning each condition to its corresponding medical department. As shown in Figure~\ref{fig:department_freq}, respiratory-related conditions dominate the dataset, introducing an inherent bias during model training. While this reflects real-world data availability, it highlights the importance of incorporating additional datasets in future work to broaden departmental coverage.

\begin{figure*}[!htbp]
    \centering
    \begin{subfigure}[b]{0.48\linewidth}
        \centering
        \includegraphics[width=\linewidth]{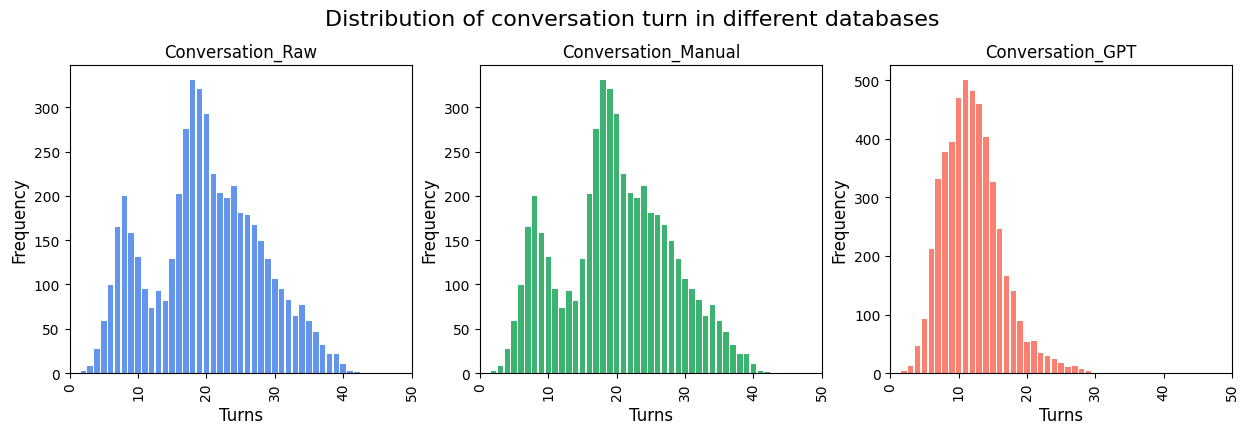}
        \caption{Distribution of conversation turns across datasets.}
        \label{fig:turn_freq}
    \end{subfigure}
    \hfill
    \begin{subfigure}[b]{0.48\linewidth}
        \centering
        \includegraphics[width=\linewidth]{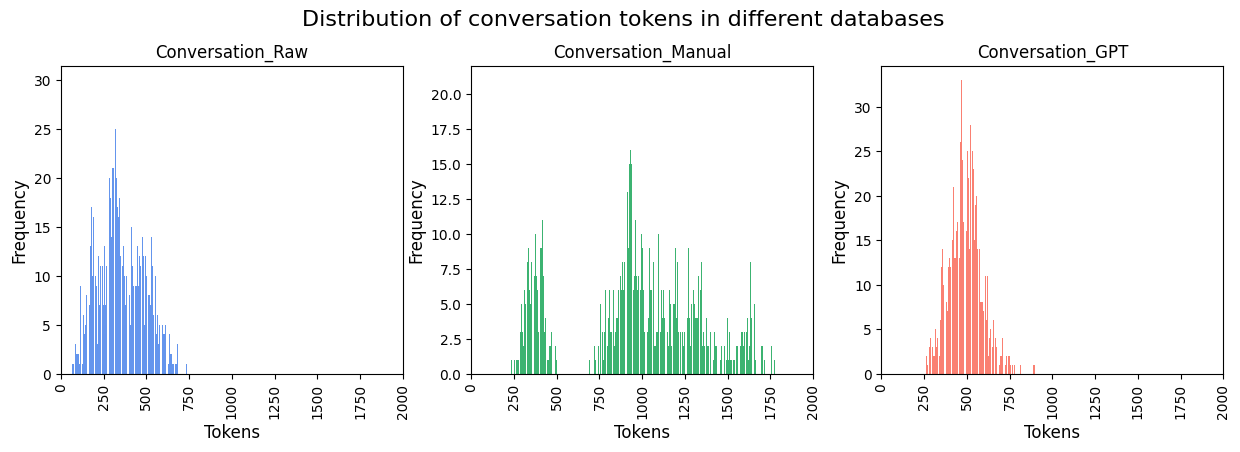}
        \caption{Token distribution across datasets.}
        \label{fig:token_freq}
    \end{subfigure}

    \vspace{0.5cm}

    \begin{subfigure}[b]{0.48\linewidth}
        \centering
        \includegraphics[width=\linewidth]{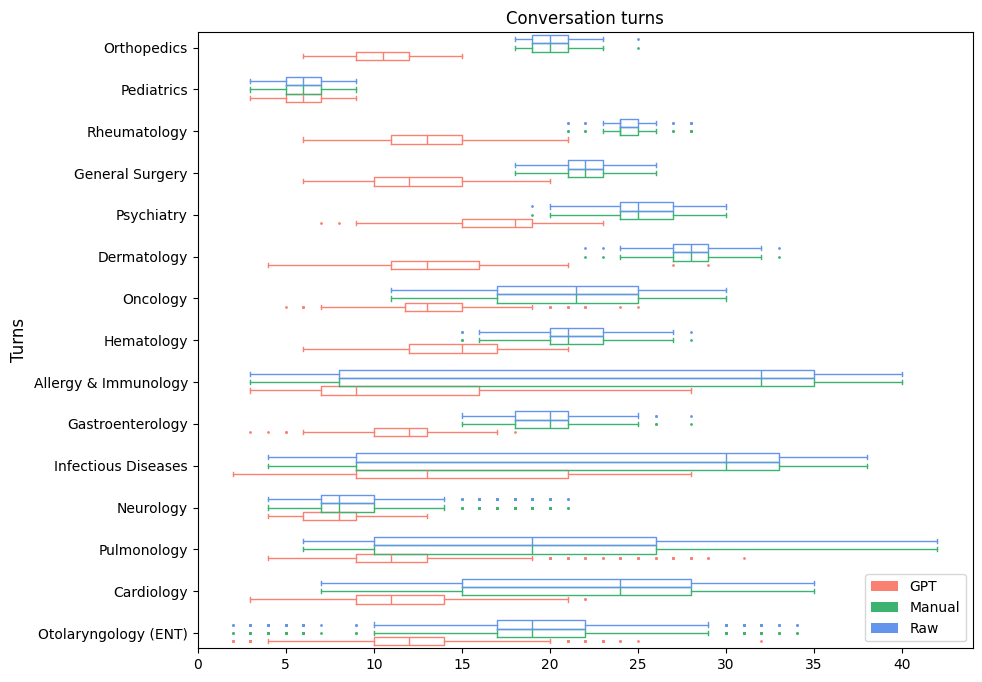}
        \caption{Turn distribution across medical departments.}
        \label{fig:turn_dept}
    \end{subfigure}
    \hfill
    \begin{subfigure}[b]{0.48\linewidth}
        \centering
        \includegraphics[width=\linewidth]{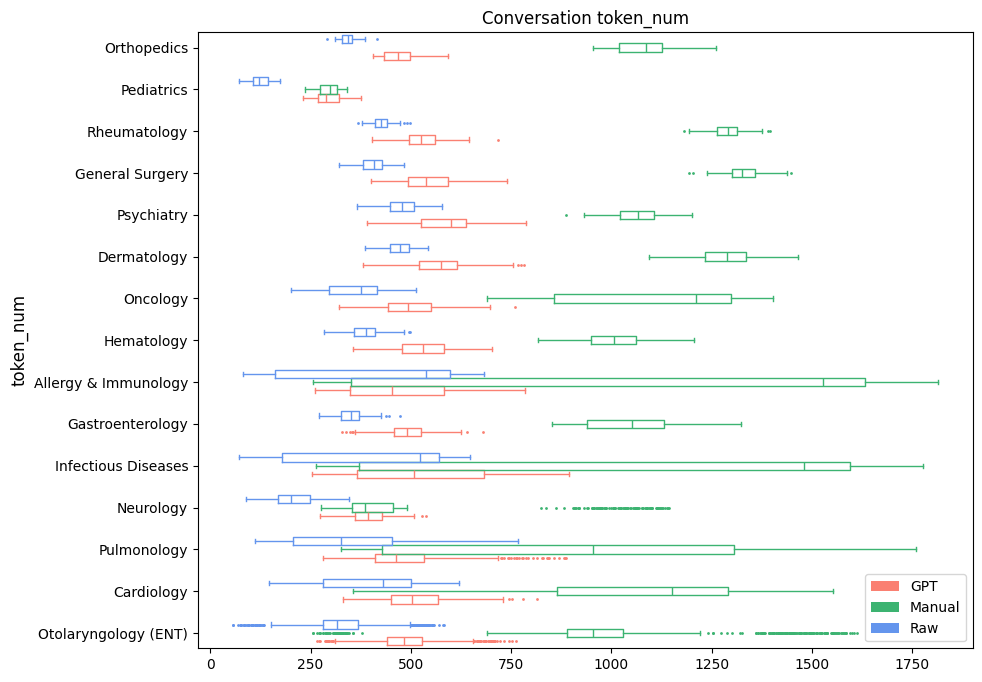}
        \caption{Token distribution across medical departments.}
        \label{fig:token_dept}
    \end{subfigure}
    
    \caption{Distributions of conversation turns and tokens across three datasets and departments}
    \label{fig:conversation_stats}
\end{figure*}

An essential challenge in LLM fine-tuning for multi-turn dialogue is the management of conversational history under input length constraints. Figure~\ref{fig:conversation_stats} analyzes the number of dialogue turns and token counts across the three datasets: \texttt{data\_5k\_ddxplus}, \texttt{data\_5k\_artificial}, and \texttt{data\_5k\_GPT}. While \texttt{data\_5k\_artificial} involves rewording original content, it maintains similar turn counts to the raw dataset, sometimes exceeding 40 turns per sample. In contrast, \texttt{data\_5k\_GPT} exhibits a smoother distribution with most conversations under 30 turns. This reduction likely reflects the more coherent structure introduced during GPT-based rewriting.

Token distribution results (Figures~\ref{fig:token_freq} and \ref{fig:token_dept}) further reveal that \texttt{data\_5k\_artificial} significantly increases token counts due to its use of expanded layperson language. These longer samples risk exceeding the 1024-token input limit, complicating training and inference. Conversely, \texttt{data\_5k\_GPT} demonstrates tighter control over token lengths, reducing context overflow and improving training stability. Additionally, the GPT rewriting process normalizes departmental variation, reducing noise and enhancing consistency in conversation structures.

\subsection{GPTScore Evaluation}

We further evaluated the quality of multi-turn conversations using GPTScore, a semantic, instruction-based metric that considers multiple conversational dimensions~\cite{fu-etal-2024-gptscore}. GPTScore provides a richer assessment than surface-level lexical metrics, making it particularly suitable for evaluating patient-facing medical dialogue.

\begin{table}[!h]
\centering
\begin{tabular}{cccc}
\toprule
\textbf{Aspect} & \textbf{data\_5k\_ddxplus} & \textbf{data\_5k\_artificial} & \textbf{data\_5k\_GPT}\\
\midrule
SPE & 70.02 & 45.47 & 19.92 \\
FLE & 81.59 & 58.05 & 62.98 \\
UND & 76.26 & 83.70 & 94.57 \\
INF & 99.90 & 100.00 & 100.00 \\
PAT & 0.10 & 0.10 & 0.00 \\
ACC & 99.60 & 100.00 & 100.00 \\
\bottomrule
\end{tabular}
\caption{GPTScore evaluation results across different datasets. }
\label{tab:gptscore_evaluation}
\end{table}

Table~\ref{tab:gptscore_evaluation} summarizes the results in six dimensions: Specificity (SPE), Flexibility (FLE), Understandability (UND), Informativeness (INF), Patience (PAT), and Accuracy (ACC). Our GPT-rewritten dataset, \texttt{data\_5k\_GPT}, achieves the highest scores in understandability and informativeness, reflecting the benefit of converting complex clinical questions into clear, accessible exchanges. The accuracy and informativeness remain consistently high across all datasets, indicating that core medical content is preserved regardless of format.

\begin{table}[!htb]
\centering
\begin{tabular}{l|cc|cc|cc}
\toprule
\textbf{Model} & \multicolumn{2}{c|}{\textbf{data\_5k\_ddxplus}} & \multicolumn{2}{c|}{\textbf{data\_5k\_artificial}} & \multicolumn{2}{c}{\textbf{data\_5k\_GPT}} \\
              & F1& Accuracy& F1   & Accuracy& F1   & Accuracy\\
\midrule
BERT& 0.773& 0.930& 0.998& 0.998& 0.996& 0.996\\
ClinicalBERT& 0.159& 0.408& 0.796& 0.964& 0.229& 0.570\\
\bottomrule
\end{tabular}
\caption{F1-Score and Accuracy of BERT and ClinicalBERT in Department Recommendation across GPT-rewritten and manually constructed dialogue datasets}
\label{tab:results}
\end{table}

Interestingly, \texttt{data\_5k\_GPT} achieves lower specificity than the raw dataset, suggesting that simplifying language may lead to more generalized expressions. However, this trade-off appears acceptable given the substantial gains in patient comprehension and coherence. The flexibility score improves over the artificial dataset, though it remains slightly behind the raw version, highlighting a potential area for prompt refinement in future work.

To further benchmark model performance in terms of classification quality, we report the F1-score and accuracy of two transformer-based models: BERT and ClinicalBERT on the department recommendation task across all three datasets. As shown in Table V, BERT achieves superior performance on \texttt{data\_5k\_artificial} and \texttt{data\_5k\_GPT}, reflecting the benefits of clearer structure and language simplification in these versions. In contrast, ClinicalBERT, despite being domain-specific, shows inconsistent results across datasets, likely due to its training objective and limited capacity to generalize over conversational inputs. These results highlight the robustness of our fine-tuned dataset design and suggest that task-specific fine-tuning can often outperform domain-specific pretraining when applied to conversational healthcare data.

\subsection{Discussions on Conversation Quality}

\begin{figure}[!h]
\centering
\includegraphics[width=\linewidth]{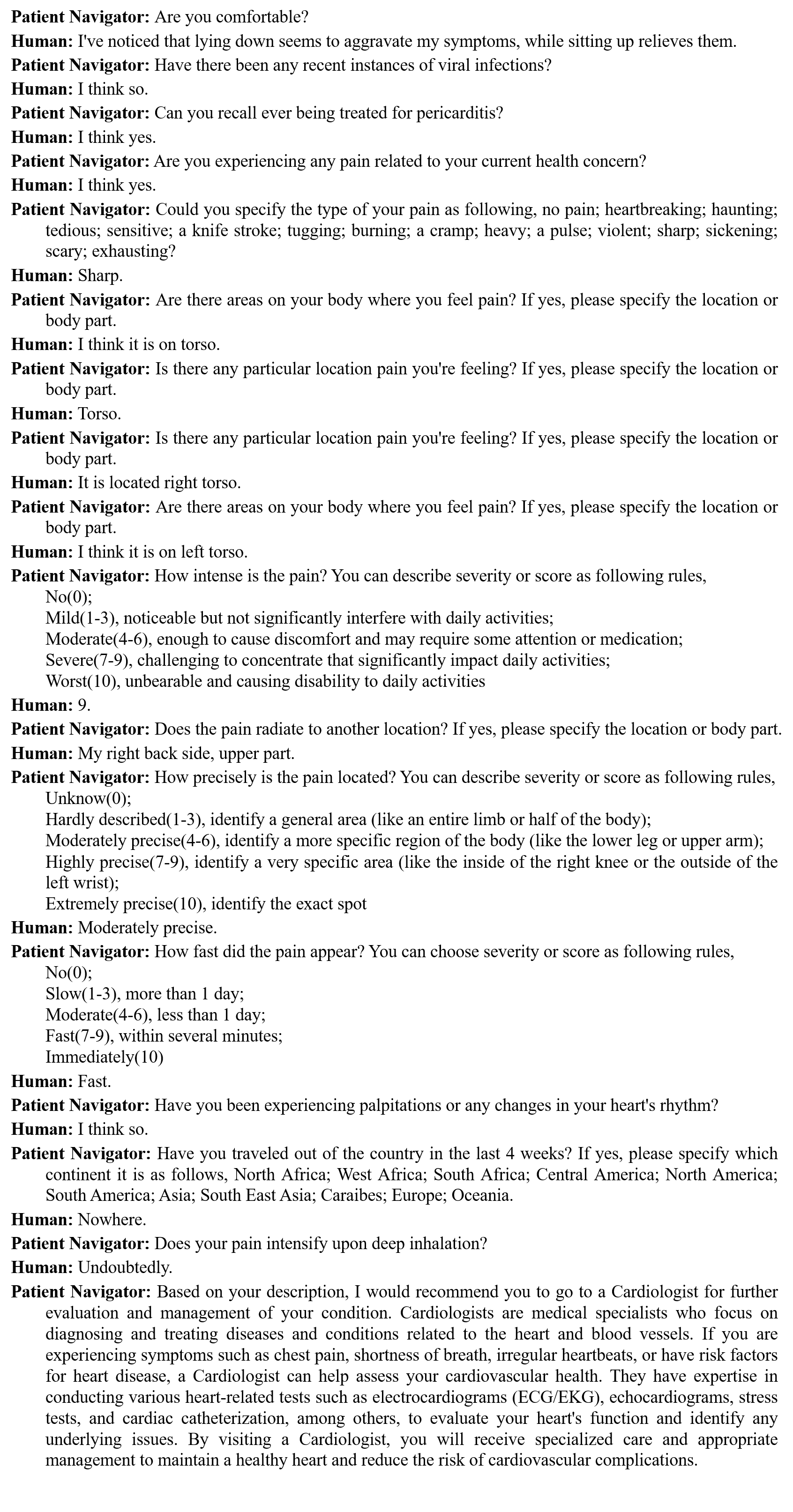}
\caption{Example conversation between a patient and the C-PATH model.}
\label{fig:conversation-example}
\end{figure}

An example conversation between C-PATH and a simulated patient is shown in Figure~\ref{fig:conversation-example}. The model engages naturally, avoiding jargon while collecting essential clinical information. This interaction style improves patient accessibility and trust, particularly for users without medical training.

The model gathers relevant details such as current symptoms, medical history, and travel patterns, allowing accurate departmental recommendations. Beyond triage, C-PATH generates structured summaries suitable for Electronic Health Record (EHR) systems, facilitating efficient physician review and reducing administrative burden.

This capability also supports healthcare systems by automating patient intake processes, reducing manual triage time, and improving throughput. C-PATH can be deployed in multiple clinical contexts, including self-service triage kiosks in hospital waiting areas, mobile health applications for at-home pre-consultation, and EHR-integrated assistant systems that streamline documentation and referral decisions for physicians. In primary care settings, the system could serve as a virtual assistant to guide patients in symptom articulation before in-person visits. In telemedicine platforms, it could support pre-screening to route patients to the appropriate specialist. These applications highlight the versatility of LLM-based medical dialogue systems and indicate their potential to improve access, efficiency, and equity in healthcare delivery.

As such, the results confirm the effectiveness of our dataset construction pipeline and fine-tuning approach. GPT-based rewriting significantly improves conversation structure and readability while preserving diagnostic value, making it an ideal foundation for instruction-tuned medical dialogue systems.

\section{Conclusions}

In this study, we introduced C-PATH, a Conversational Patient Navigator powered by a fine-tuned Large Language Model (LLM) that assists users in symptom recognition and triage to appropriate medical departments. Using a multi-stage training pipeline that comprises acquisition of medical knowledge, conversational alignment, and summarization, we demonstrated that LLMs can be adapted to produce safe, understandable, and context-aware medical dialogues. C-PATH enables laypeople to communicate symptoms naturally, ask follow-up questions, and receive accurate referral suggestions, enhancing both patient experience and healthcare workflow efficiency.

A central innovation of this work is our GPT-based rewriting pipeline that transforms structured clinical data from DDXPlus into accessible, multi-turn conversations. This method bridges the gap between domain-specific medical knowledge and patient-friendly communication, significantly improving the clarity of dialogue and the usability of the model. Additionally, we introduced a lightweight dialogue history management strategy combining sliding window pruning and optional summarization, allowing for scalable multi-turn interactions within LLM context limits.

Our evaluations showed that C-PATH outperforms baselines in both conversation quality and triage accuracy, and the GPTScore metrics confirm improvements in understandability, informativeness, and conversational flow. These capabilities make C-PATH well suited for deployment in digital health settings such as patient self-assessment tools or intake assistants, with the potential to reduce provider burden and accelerate access to care.

However, several limitations remain. Logical reasoning challenges inherent to LLMs, such as the ``Reversal Curse"~\cite{berglund2024the}, restrict their ability to generalize in nuanced medical scenarios. Furthermore, hallucination remains a concern in patient-facing contexts, especially when models confidently present incorrect information due to training artifacts~\cite{Ji2023}. In addition, the training datasets derived from DDXPlus exhibit imbalances in department representation, with certain specialties such as respiratory and general medicine being overrepresented. This skew may reduce the model’s ability to generalize recommendations across underrepresented or rare medical departments. We also note the importance of reinforcement learning from human feedback (RLHF) to align model behavior with clinical standards, which is not yet integrated in this iteration of C-PATH.

Future work will explore the incorporation of RLHF with clinical expert supervision, expanding the training data to include more diverse medical domains, and conducting real-world user studies with patients and clinicians. We also aim to integrate our model into mobile health platforms, enabling greater accessibility and continuous improvement through conversational data feedback. C-PATH represents a step toward scalable, trustworthy AI assistance in healthcare and lays the foundation for future patient-centered dialogue systems.

\section*{Acknowledgements}
This work was funded by Fundação para a Ciência e a Tecnologia (UIDB/00124/2020, UIDP/00124/2020 and Social Sciences DataLab - PINFRA/22209/2016), POR Lisboa and POR Norte (Social Sciences DataLab, PINFRA/22209/2016) and NOVA LINCS (grant UIDB/04516/2020) and Carnegie Mellon
Portugal Program (CMU/TIC/0016/2021). This work was supported in part by Oracle Cloud credits and related resources provided by Oracle for Research. We also acknowledge the use of ChatGPT, an AI language model by OpenAI, for assistance with grammar, LaTeX formatting, and general English writing improvements throughout the manuscript.


\bibliographystyle{plain}
\bibliography{Manuscript}

\end{document}